\documentclass{sjtudenglab}
\usepackage{amsmath}
\usepackage{enumerate} 
\usepackage{algorithm}
\usepackage{algpseudocode}
\usepackage{amsfonts}
\usepackage{amsthm}
\usepackage{newtxtt}
\usepackage{diagbox} 
\usepackage{colortbl}
\usepackage{amssymb}
\usepackage{xspace}
\usepackage{wrapfig}
\usepackage{adjustbox}
\usepackage{tabularx}
\usepackage{booktabs}
\usepackage{mathtools}
\usepackage{tikz}
\usepackage{enumitem}
\usepackage{silence}
\usepackage{dsfont}
\usepackage[table]{xcolor}
\usepackage[dvipsnames]{xcolor}
\usepackage{multirow}
\usepackage{makecell}
\usepackage{xfakebold}
\usepackage{natbib}
\usepackage{siunitx}
\usepackage{cleveref}
\usepackage{thmtools}

\usepackage{booktabs}
\usepackage{multirow}
\usepackage{makecell}
\usepackage{caption}

\usepackage{colortbl}
\usepackage{pifont}
\usepackage{graphicx} 
\usepackage{amsmath}
\usepackage{amssymb}
\usepackage{float}

\usepackage{amsmath,amsfonts,bm}

\def\eqref#1{equation~\ref{#1}}

\def\1{\bm{1}}

\DeclareMathAlphabet{\mathsfit}{\encodingdefault}{\sfdefault}{m}{sl}
\SetMathAlphabet{\mathsfit}{bold}{\encodingdefault}{\sfdefault}{bx}{n}

\usepackage{parskip}

\usepackage{natbib}
\usepackage{amsfonts,bm}

\usepackage{amsmath, amssymb,mathrsfs, amsthm, mathtools}
\usepackage{xspace}
\usepackage{enumitem}
\usepackage{lipsum}
\usepackage{xpatch}
\usepackage{mathtools}
\usepackage{dsfont}
\usepackage{pgf,tikz}
\usetikzlibrary{positioning,matrix,patterns}
\usepackage{caption}
\usepackage{graphicx}
\usepackage{makecell}
\usepackage{multirow}
\usepackage[mathscr]{euscript}
\definecolor{hanblue}{rgb}{0.27, 0.42, 0.81}
\definecolor{deepred}{HTML}{900C3F}
\definecolor{deepgreen}{HTML}{2F6960}
\hypersetup{
    colorlinks=true,
    linkcolor=hanblue,
    urlcolor=hanblue,
    citecolor=hanblue,
    anchorcolor=hanblue}

\declaretheoremstyle[
  headfont=\sffamily\bfseries,
]{sansserif}

\theoremstyle{sansserif}

\theoremstyle{definition}

\theoremstyle{sansserif}

\theoremstyle{remark}

\DeclarePairedDelimiter\abs{\lvert}{\rvert}%
\DeclarePairedDelimiter\norm{\lVert}{\rVert}%

\makeatletter
\let\oldabs\abs
\def\abs{\@ifstar{\oldabs}{\oldabs*}}
\let\oldnorm\norm
\def\norm{\@ifstar{\oldnorm}{\oldnorm*}}
\makeatother

\definecolor{textgray}{HTML}{6E6E73}
\usetikzlibrary{positioning, calc}
\usetikzlibrary{decorations.pathmorphing}

\makeatletter
\patchcmd{\wrong@fontshape}{\@gobbletwo}{}{}{}
\makeatother
\numberwithin{equation}{section} 
\tcbuselibrary{minted}
\setminted[python]{
  linenos,
  breaklines,
  fontsize=\footnotesize,
  xleftmargin=2em
}

\makeatletter
\AtBeginDocument{
  \urlstyle{sf}
  
}
\makeatother

\floatstyle{plain}
\newfloat{algorithmgroup}{tbp}{loag}
\floatname{algorithmgroup}{Algorithm Group}

\renewcommand{\eqref}[1]{\textup{(\ref{#1})}}

\newcommand{\lstheading}[1]{{\normalfont\bfseries\textcolor[RGB]{144,162,179}{#1}}}
\newcommand{\subtasklabel}[1]{{\normalfont\bfseries\textcolor[RGB]{144,162,179}{s\ensuremath{_{#1}}:}}}

\lstdefinestyle{promptstyle}{
    basicstyle=\ttfamily\scriptsize,
    breaklines=true,
    breakatwhitespace=false,
    frame=none,
    columns=fullflexible,
    breakindent=0pt,
    breakautoindent=false,
    aboveskip=0pt,
    belowskip=0pt,
    escapeinside={(*@}{@*)}
}

\definecolor{light}{RGB}{125, 125, 125}
\crefname{tcb@cnt@pbox}{code}{code}
\Crefname{tcb@cnt@pbox}{Code}{Code}
\crefname{assumption}{assumption}{assumption}
\Crefname{assumption}{Assumption}{Assumptions}

\newtcolorbox[auto counter]{pbox}[2][]{
  colback=white,
  title=Code~\thetcbcounter: #2,
  #1,fonttitle=\sffamily,
  fontupper=\sffamily,
  arc=2pt,
  colframe=bgcolor,
  coltitle=fgcolor,
  colbacktitle=bgcolor,
  toptitle=0.25cm,
  bottomtitle=0.125cm
}

\makeatletter
\newcommand\applefootnote[1]{%
  \begingroup
  \renewcommand\thefootnote{}%
  \renewcommand\@makefntext[1]{\noindent##1}%
  \footnote{#1}%
  \addtocounter{footnote}{-1}%
  \endgroup
}
\makeatother

\definecolor{cverbbg}{gray}{0.90}

\title{World-Language-Action Model for Unified World Modeling, \\ Language Reasoning, and Action Synthesis}

\renewcommand{\authorlist}{%
  {\sffamily \sbseries
  Yi Yang       $^{1,2,\dagger}$,
  Zhihong Liu   $^{1,\dagger}$,
  Siqi Kou      $^{1,\dagger}$,
  Yiyang Chen   $^{1}$,
  Yanzhe Hu     $^{3}$,
  Jianbo Zhou   $^{4}$,
  \\
  Boyuan Zhao   $^{5}$,
  Zhijie Wei    $^{6}$,
  Xiao Xia      $^{1}$,
  Xueqi Li      $^{2,7}$,
  Pengfei Liu   $^{1,2}$,
  Zhijie Deng   $^{1,\ddagger}$
  
  }%
}

\affiliation[1]{SJTU}
\affiliation[2]{SII}
\affiliation[3]{HUST}
\affiliation[4]{SCUT}
\affiliation[5]{ECUST}
\affiliation[6]{SHU}
\affiliation[7]{NJUPT}

\vspace{111em}

\contribution[\dagger]{Equal Contribution}
\contribution[\ddagger]{Corresponding Author}

\abstract{
We propose world-language-action (WLA) models as a new class of embodied foundation models. WLA takes textual instructions, images, and robot states as inputs to jointly predict textual subtasks, subgoal images, and robot actions, conjoining the \emph{world modeling interface} to learn from extensive egocentric videos as in the world-action model (WAM) and the \emph{language reasoning} capacities to solve complex long-horizon tasks as in vision-language-action (VLA) models. At the core of WLA lies an \emph{autoregressive (AR)} Transformer backbone, instead of a bidirectional diffusion Transformer as in WAMs, to predict the \emph{next state}, comprising the \emph{semantic-level} textual intention and complementary \emph{fine-grained} physical dynamics. The physical dynamics are supervised by the world modeling objective based on a dedicated World Expert, and are leveraged to ease the characterization of the state-action correlation for the Action Expert. WLA leverages meta-queries to make the world prediction \emph{implicitly} impact the action generation so that the former can be disabled during inference. The world prediction can also be activated to enable test-time scaling for improved robot control. Our WLA-0 prototype, with 2B active parameters, achieves 40 ms per inference on an NVIDIA RTX 5090. Evaluations across simulated and real-world environments demonstrate that WLA-0 achieves state-of-the-art multi-task and long-horizon learning abilities, e.g., 92.94\% success rate on RoboTwin2.0 Clean and 56.5\% success rate on RMBench. WLA-0 also holds the promise to learn novel tasks directly from \emph{cross-embodiment robot videos} without action annotations.

}

\date{\sffamily\today}
\metadata[Code]{\url{{https://github.com/SJTU-DENG-Lab/WLA}}} 

\AtBeginDocument{ 
  \fancyheadoffset[R]{-0.3cm}
  \fancyhead[R]{\raisebox{9pt}{\includegraphics[height=35pt]{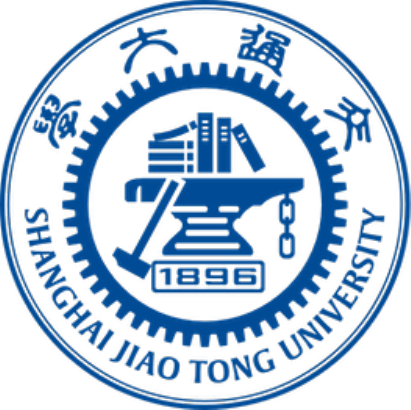}}}
}

\begin{document}

\maketitle

\section{Introduction}

World models (WMs)~\cite{ha2018world, agarwal2025cosmos, brooks2024video, bruce2024genie} aim to model physical dynamics and underpin physical AI. 
Recently, world-action models (WAMs)~\cite{ye2026world, pai2025mimic, kim2026cosmos} have emerged as a compelling paradigm that integrates WMs for embodied control.
The world modeling interface enables WAMs to benefit from large-scale pretraining on egocentric videos. 
The prediction of physical dynamics provides strong future-state priors that facilitate effective action prediction. 
However, current methods focus almost exclusively on predicting the \emph{next visual state}, burdening models with low-level details and restricting their capacity for semantic reasoning and extrapolation.

To bridge this gap, our key insight is that the next state should comprise both high-level textual intention and low-level physical dynamics.
Specifically, the former offers a compact, highly generalizable abstract representation of future states, which can be readily obtained given the prevalence of large language models (LLMs)~\cite{brown2020language}. 
The latter serves as a bridge between high-level intention and fine-grained motion control, but it differs from the high-resolution visual state in that it only describes the transitions between such states.

We propose world-language-action (WLA) models, a new family of embodied foundation models, to connect such next-state prediction to action synthesis. 
WLA adopts an autoregressive (AR) Transformer capable of text generation as the backbone, which stands in stark contrast to existing WAMs built upon bidirectional diffusion Transformers (DiT)~\cite{peebles2023scalable,wan2025wan}. 
In practice, WLA confines the high-level intention to textual subtasks decomposed from the original instructions, 
and opts to inherit the language modeling abilities and context management schemes of existing vision-language models (VLMs)~\cite{bai2023qwen, liu2023visual}. 
Compared with vision-language-action (VLA) models~\cite{zitkovich2023rt, kim2024openvla}, WLA adopts the high-level intention to guide the generation of both physical dynamics and action prediction — whereas VLA rarely does so for action prediction. 
Accordingly, WLA can exploit heterogeneous data, including cross-embodiment robot videos, with or without action annotations.

\begin{figure}[t]
    \centering
    \includegraphics[width=1\linewidth]{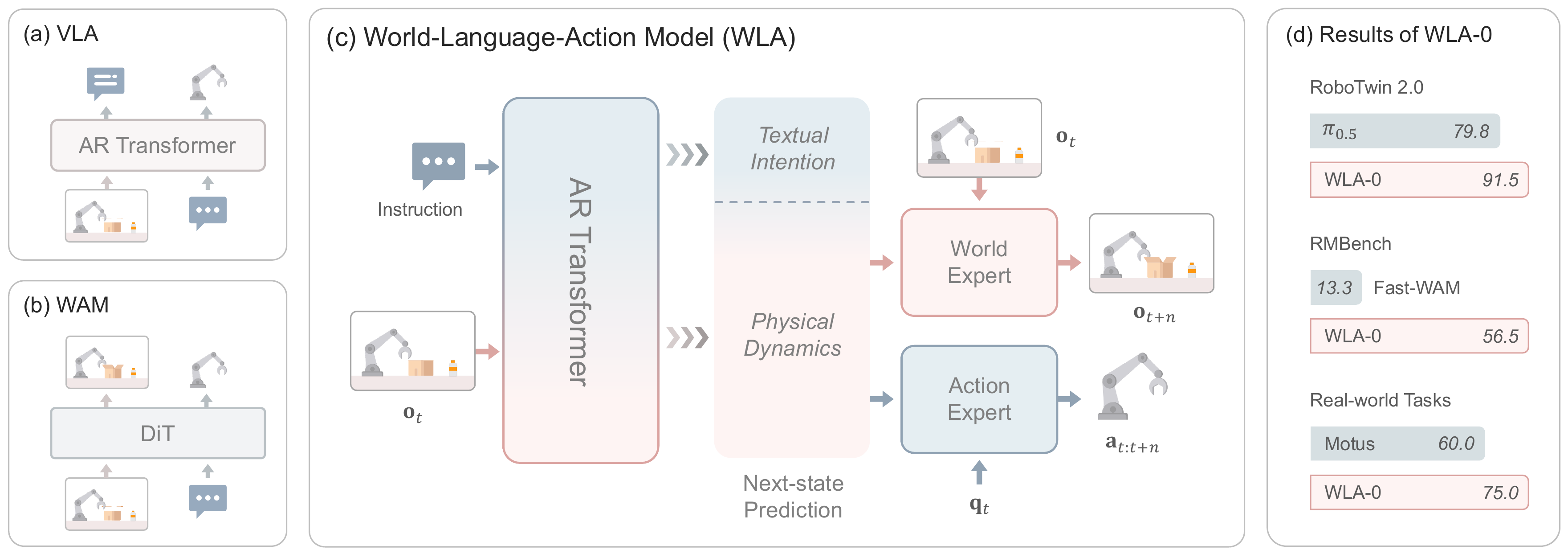}
    \caption{
    (a) VLA architecture.
    (b) WAM architecture.
    (c) WLA uses an autoregressive (AR) Transformer backbone to predict the next state from two complementary representations: high-level textual intention and low-level physical dynamics.
    (d) WLA-0 achieves strong performance with only 2B active parameters and no embodied pretraining.
    }
    \label{fig:head}
\end{figure}

Enabling the AR backbone to predict the low-level physical dynamics can be non-trivial due to the absence of ground truth.
WLA addresses this by introducing a dedicated World Expert to predict the subsequent visual state based on the current state as well as the physical dynamics yielded by the backbone. 
Such a world modeling objective lifts the burden of visual detail prediction to the World Expert, allowing the backbone to predict only the core information driving the transition of visual states, which is also known as the \emph{latent action}. 
Unlike existing methods with explicit {latent action} learning~\cite{ye2025latent,bi2025motus,chen2025moto,bu2025univla}, our framework is trained in an end-to-end manner rather than following a two-stage pipeline. 
The implementation is based on a simple meta-query~\cite{pan2025transfer} architecture on top of the AR backbone.
The outputs of meta-queries act as conditioning signals for the World Expert, and also guide the Action Expert to produce executable actions.

We have identified several crucial design insights to preserve the efficiency of WLA. We find that equipping the World Expert with a lightweight diffusion Transformer such as SANA-600M~\cite{xie2024sana} and predicting only static future visual frames rather than full video clips suffices to capture valid physical dynamics. 
Because the world prediction influences the action generation via \emph{implicit} parameter updates rather than explicit conditional modeling, the World Expert can be disabled during inference. 
Our first version, WLA-0, achieves $\sim$40 ms inference latency on an RTX 5090, enabling real-time adaptation in dynamic environments.

Extensive experiments show that WLA-0 substantially improves generalization, inference efficiency, and long-horizon task performance.
Despite having only 2B running parameters and no pretraining, it matches leading WAMs on simulation benchmarks~\cite{chen2025robotwin, liu2023libero}, with test-time scaling yielding further performance gains.
Real-world evaluations demonstrate robustness in dynamic and out-of-distribution (OOD) settings; on \textit{Stack Cup}, WLA-0 halves the completion time of the baseline WAM, highlighting its suitability for latency-sensitive control.
On RMBench~\cite{chen2026rmbench}, a long-horizon, memory-dependent benchmark, WLA-0 sets a new state-of-the-art (SOTA) by leveraging language-based planning, memory use, and error correction, nearly doubling the performance of the best baseline.
Furthermore, WLA-0 can learn new tasks from cross-embodiment videos without action annotations, 
demonstrating promising steerability and cross-embodiment generalization.
\section{Related Work}
\label{sec:related_work}

\subsection{World Modeling for Policy Learning}
Recently, a growing body of work has incorporated world modeling with embodied control to enhance policy learning, typically by predicting future visual states~\cite{wu2024unleashing, luo2025being, cheang2024gr}.
Early methods generally rely on explicit inverse dynamics to infer actions from current observations and predicted future states~\cite{du2023learning, ko2024learning, feng2025vidar}, whereas later approaches treat visual prediction as an intermediate chain-of-thought (CoT)~\cite{wei2022chain} reasoning step for action generation~\cite{zhao2025cot, wang2025unified, zhang2026dreamvla}. More recent methods further exploit the internal representations of video diffusion models to guide action prediction through implicit inverse dynamics~\cite{hu2024video, pai2025mimic}.
The convergence of world modeling and action generation has spurred the emergence of World Action Models (WAMs). Some WAMs build on pretrained video generation models (VGMs), leveraging their learned physical priors to improve policy learning~\cite{li2026causal, kim2026cosmos}. Others adopt Mixture-of-Transformers (MoT) architectures~\cite{liang2024mixture} to jointly model task understanding, video prediction, and robot control~\cite{bi2025motus, ma2026dit4dit, yuan2026fast}. By training on large-scale web and egocentric video data, WAMs acquire rich interaction priors that enhance downstream generalization and data efficiency~\cite{ye2026world}. 
Nevertheless, most WAMs rely on VGM backbones lacking language generation capabilities, limiting high-level planning and reasoning.
WLA addresses this with a VLM backbone that preserves native language functionality for language-guided reasoning and planning.

\subsection{Language-Guided Embodied Control}

Language provides a natural interface for instruction following, high-level planning, and behavioral steering in embodied agents. 
By leveraging the rich perceptual and linguistic representations of pretrained VLMs~\cite{driess2023palm, karamcheti2024prismatic, beyer2024paligemma}, VLA models enable robots to follow human commands tightly and extend beyond reactive control policies~\cite{zitkovich2023rt, zhou2025chatvla}.
Subsequent work further shows that language bridges the gap between high-level goals and low-level control in long-horizon tasks: through CoT reasoning and hierarchical planning, models decompose complex instructions into structured subtasks and translate them into executable action sequences~\cite{zawalski2024robotic, mu2023embodiedgpt, intelligence2025pi_}.
More recent advances frame language as a flexible conditioning mechanism for steerable control, improving generalization to unseen environments and compositional tasks~\cite{intelligence2026pi}.
Despite recent advances, existing methods still struggle to capture physical dynamics, while the lack of visual supervision leads to insufficient training signals, thereby constraining the model’s capabilities~\cite{li2025drivevla}.
Instead, WLA unifies world modeling, language-guided planning, and action prediction within a single framework.
\section{Methodology}
\label{sec:methodology}

This paper aims to develop a unified foundation model for physical AI that maps multimodal inputs (images, text, and robot states) to multimodal outputs (images, text, and robot actions).
This formulation supports heterogeneous supervision, including image-text pairs, robot demonstrations, and egocentric videos, thereby combining the strengths of WAMs and VLAs. 
At each time step $t$, the model processes the current observation $\mathbf{o}_t$, a historical observation $\mathbf{o}_{t-h}$, the proprioceptive state $\mathbf{q}_t$, and the instruction $\ell$, predicting an $n$-step action chunk $\mathbf{a}_{t:t+n}$, which is preceded by textual intention $\hat{\ell}$ and the future visual state $\mathbf{o}_{t+n}$.
Executing $\mathbf{a}_{t:t+n}$ advances the environment toward $\mathbf{o}_{t+n}$ in a receding-horizon manner, repeating until task completion.

\subsection{World-Language-Action Models} 
\begin{figure*}[t]
    \centering
    \includegraphics[width=1\linewidth]{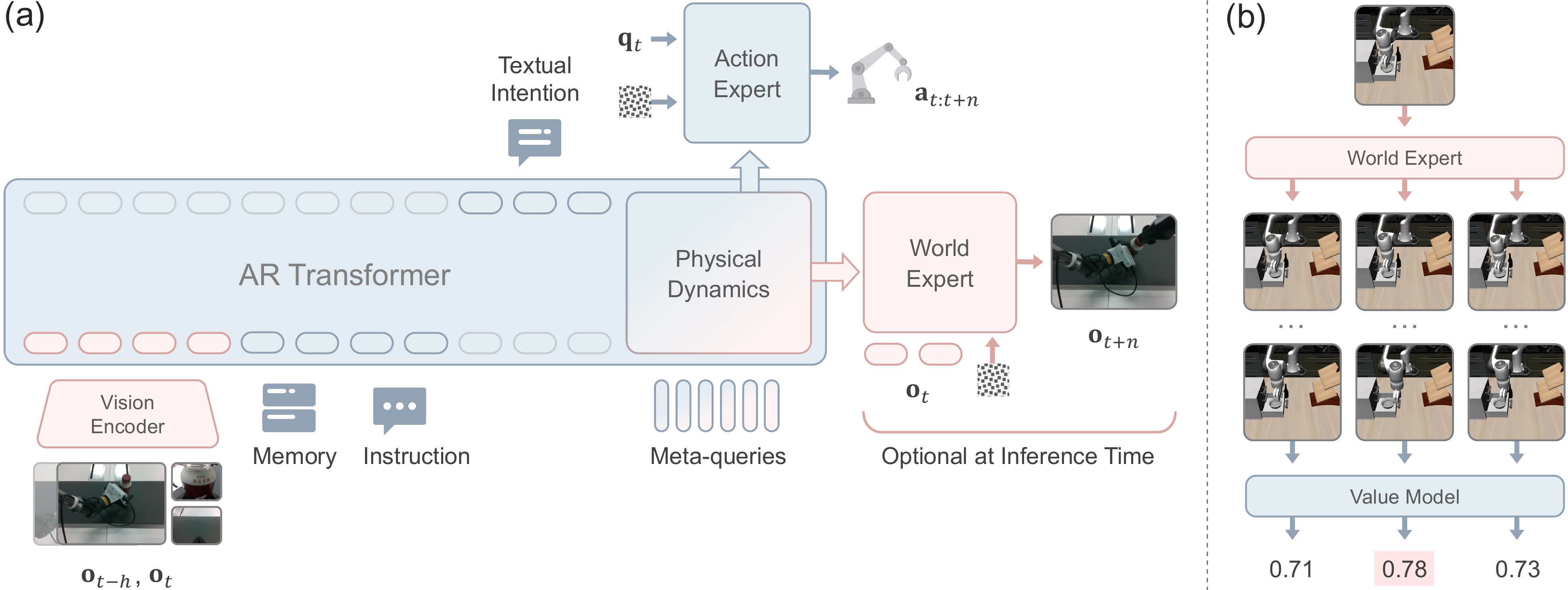}
    \caption{
    (a) WLA comprises three components: an AR Transformer backbone for next-state prediction, a World Expert for forecasting future observations, and an Action Expert for action generation.
    (b) In test-time scaling mode, WLA executes the action chunk with the highest predicted value.
    }
    \label{fig:architecture}
\end{figure*}

As shown in Figure~\ref{fig:architecture}, WLA employs an AR Transformer backbone to predict the next state through two complementary representations: high-level textual intention and low-level physical dynamics. 
This is in stark contrast to existing WAMs, which leverage bidirectional DiTs~\cite{peebles2023scalable} to predict purely visual states.
The textual intention stream provides a semantic blueprint of state evolution, offering global guidance for robot behavior.
In parallel, the physical dynamics capture state transitions, effectively grounding high-level directives in low-level motion patterns.
We expand the details below.

\textbf{Textual Intention Learning.}
For robot control, high-level intentions are naturally textual subtasks decomposed from the original user instructions, offering compact and faithful descriptions of future robot actions. 
In this sense, WLA initializes the backbone $f$ with a pretrained VLM to leverage its rich contextual representations. 
For training, we first construct a sequence of intermediate subtasks $\hat{\mathcal{L}}=\{\hat{\ell}_1, \hat{\ell}_2, \dots, \hat{\ell}_N\}$ corresponding to $\ell$, where each subtask $\hat{\ell}_k$ is associated with a temporal segment $[s_k,e_k]$.
WLA is trained to predict a contiguous subtask window
$\mathcal{S}_t=\{\hat{\ell}_{k_t},\ldots,\hat{\ell}_{k_{t+n}}\} \subseteq \hat{\mathcal{L}}$
that spans the upcoming action horizon $[t,t+n]$ (i.e., $s_{k_t} \leq t$ and $e_{k_{t+n}}\geq t+n$):
\begin{equation}
  \mathcal{S}_t = f(\mathbf{o}_{t-h}, \mathbf{o}_t, \ell, \mathcal{M}).
  \label{eq:ell}
\end{equation}
$\mathcal{M}$ denotes a memory buffer to serve as historical context for subsequent predictions for long-horizon tasks. 
It will be updated by $ \mathcal{M} \leftarrow \mathcal{M}\oplus[\hat{\ell}_{k_t},\ldots,\hat{\ell}_{k_{t+n}-1}]$ recursively.

\textbf{Physical Dynamics Modeling.}
To model the physical dynamics with the AR backbone, we employ a World Expert $f_{\mathrm{wm}}$ and optimize it with a world modeling objective.
Specifically, we append a set of meta-queries $\mathbf{Q}$~\cite{pan2025transfer} to the context of the AR backbone, allowing them to aggregate contextual information through causal attention, and use the outputs to define the physical dynamics $\mathbf{h}_t$, i.e.,
\begin{equation}
  \mathbf{h}_t = f(\mathbf{o}_{t-h}, \mathbf{o}_t, \ell, \mathcal{M}, \mathcal{S}_t, \mathbf{Q}).
  \label{eq:ht}
\end{equation}

The World Expert $f_{\mathrm{wm}}$ then accepts both $\mathbf{h}_t$ and the representation of the original state $\mathbf{o}_t$ (e.g., yielded by the vision encoder of the used VLM) to predict the future visual state $\mathbf{o}_{t+n}$:
\begin{equation}
  \mathbf{o}_{t+n}=f_{\mathrm{wm}}(\mathbf{h}_t, \mathbf{o}_t).
  \label{eq:wm_expert_expert}
\end{equation}
WLA keeps $\mathbf{h}_t$ compact so that it captures only the core visual transitions, while leaving fine-grained detail prediction to $f_{\mathrm{wm}}$.
Rather than directly predicting $\mathbf{o}_{t+n}$, $f_{\mathrm{wm}}$ predicts its VAE~\cite{kingma2013auto} feature representation.
We use VAE features instead of semantic features such as DINO~\cite{caron2021emerging} or JEPA~\cite{bardes2024revisiting}, since the physical dynamics in our framework are already modeled at the semantic level and require no additional semantic inductive bias.
We further design $f_{\mathrm{wm}}$ to predict only the target frame $\mathbf{o}_{t+n}$ rather than the full video clip $\mathbf{o}_{t:t+n}$. 
This is motivated by experimental evidence that full-clip supervision slows training without improving performance (Appendix~\ref{app:simulation benchmarks}).
The model can optionally predict depth maps alongside $\mathbf{o}_{t+n}$ to enhance spatial supervision.

$\mathbf{h}_t$ can be interpreted as a \emph{latent action}~\cite{ye2025latent,bi2025motus,chen2025moto}. The key distinction from prior work is that existing methods adopt pretrained action quantizers, while our framework is trained end-to-end and thus avoids suboptimal optimization. 
Intuitively, $\mathbf{h}_t$ contains the minimal sufficient information for steering the Action Expert $f_{\mathrm{act}}$ to generate explicit actions. 
I.e., there is:
\begin{equation}
  \mathbf{a}_{t:t+n}=f_{\mathrm{act}}(\mathbf{h}_t, \mathbf{q}_t),
  \label{eq:wm_act_expert}
\end{equation}

Overall, the world modeling objective guides action generation via shared parameter learning during training, rather than requiring the action model to condition on explicitly predicted future images at test time. 
Such an \emph{implicit} paradigm enables $f_{\mathrm{wm}}$ to be entirely discarded during inference. 
This departs from the traditional ``image-then-act'' WAM, significantly reducing test-time latency.

\textbf{Training Objective.} 
The model is jointly trained with a cross-entropy loss 
$\mathcal{L}_{\mathrm{lang}}$ for subtask generation and two flow-matching losses, 
$\mathcal{L}_{\mathrm{wm}}$ and $\mathcal{L}_{\mathrm{act}}$, for world modeling and action prediction:
\begin{equation}
 \mathcal{L}=\mathcal{L}_{\mathrm{act}}+\alpha \mathcal{L}_{\mathrm{wm}}+\beta \mathcal{L}_{\mathrm{lang}},
  \label{eq:loss}
\end{equation}
where $\alpha$ and $\beta$ weight the auxiliary world-modeling and language losses, respectively.

\subsection{Inference Optimization}

We additionally contribute a novel test-time scaling (TTS) paradigm for robot control. 
We describe the original inference scheme (i.e., efficient mode) and the TTS mode of WLA below.

\textbf{Efficient Mode.} By default, WLA disables the World Expert during inference, as described above. We further improve deployment efficiency through several acceleration techniques in Appendix~\ref{app:acceleration techniques}. Together, these designs allow WLA to achieve real-time action prediction with $\sim$40 ms inference latency on an RTX 5090.

\textbf{TTS Mode.} When more computation is affordable, WLA can switch to the test-time scaling mode to further improve action prediction, as shown in Fig.~\ref{fig:architecture} (b). Given the current observation, we sample $K$ candidate action chunks by varying the random seed. For each candidate $k$, the World Expert predicts the corresponding future static frame $\hat{\mathbf{o}}_{t+n}^{(k)}$, which represents the imagined visual state after executing the candidate action chunk $\hat{\mathbf{a}}_{t:t+n}^{(k)}$. A value model then scores these imagined future states, and WLA executes the action chunk with the highest predicted value. This allows WLA to reject potentially failing trajectories in the imagined space before they affect the real environment. The imagination horizon can be extended by autoregressively using the predicted future frame as the next input. 

The value model is trained using rollouts from the fine-tuned WLA-0. Each rollout contains the task instruction $\ell$, a sequence of World-Expert-predicted future frames $\{\hat{\mathbf{o}}_{in}\}_{i=1}^{\lfloor T/n \rfloor}$, and a binary success indicator $y \in \{0,1\}$, where $T$ is the episode length. For a predicted future frame at time step $t$, we assign the discounted value label as $v_t = y \cdot \gamma^{T-t}$, where $\gamma < 1$ is the discount factor. The value model is then trained to estimate this label from the task instruction and the predicted future frame.
\section{Experiments}
\label{sec:experiments}

\subsection{Implementation Details}
We instantiate WLA-0 (3.4B total parameters) using RynnBrain-2B~\cite{dang2026rynnbrain} (2.1B) as the backbone, SANA-600M~\cite{xie2024sana} (900M including the VAE) as the World Expert, and a flow-matching head~\cite{community2026starvla} (390M) as the Action Expert. 
Each expert has 28 layers. 
We set the number of meta-queries to 64, the action chunk size to 8 for the LIBERO~\cite{liu2023libero} benchmark and 32 elsewhere. 
For efficient distributed training, we leverage DeepSpeed~\cite{rasley2020deepspeed} and optimize the model using AdamW~\cite{loshchilov2017decoupled} (weight decay $1\times10^{-8}$, gradient clipping 1.0). The learning rate follows a cosine schedule (base LR $5\times10^{-5}$, min LR $5\times10^{-6}$ with 1,000 warm-up steps). The loss weights in Eq.~\ref{eq:loss} are set to $\alpha=0.1$ and $\beta=0.005$.


\subsection{Evaluation in Simulation Environment}

\begin{table*}[t]

\caption{
    \textbf{Comparison on RoboTwin 2.0 and LIBERO benchmark.}
    WLA-0 achieves performance comparable to state-of-the-art WAM baselines, while activating only 2B parameters during inference and requiring no embodied pretraining.
    \textbf{Bold} and \underline{\textit{Italics}} indicate the best and second-best results, respectively.
    $-\mathcal{L}_\mathrm{wm}$ denotes the variant without the World Expert loss, and $+$\textit{TTS} denotes test-time scaling mode.
}
\renewcommand{\arraystretch}{1.2}
\centering
\small
\label{tab:robotwin_and_libero}

\begin{tabular}{l|c|c|cc|ccccc}
\toprule[0.5mm]
\multirow{2}{*}{Method} 
& \multirow{2}{*}{\makecell{Active\\Params}} 
& \multirow{2}{*}{\makecell{Embodied\\Pretraining}} 
& \multicolumn{2}{c|}{RoboTwin 2.0} 
& \multicolumn{5}{c}{LIBERO} \\
& & & \textit{Clean} & \textit{Rand.} & \textit{Spatial} & \textit{Object} & \textit{Goal} & \textit{Long} & \textit{\textbf{Avg.}} \\
\midrule

$\pi_0$~\cite{black2024pi_0}    
    & \textit{3B}    
    & \ding{51}    
    & 65.92 & 58.40 
    & 96.8 & 98.8 & 95.8 & 85.2 & 94.2 \\

$\pi_{0.5}$~\cite{intelligence2025pi_}    
    & \textit{3B}    
    & \ding{51}    
    & 82.74 & 76.76 
    & 98.8 & 98.2 & \underline{\textit{98.0}} & 92.4 & 96.9 \\

Motus~\cite{bi2025motus}    
    & \textit{8B}    
    & \ding{51}    
    & 88.66 & 87.02 
    & 96.8 & \underline{\textit{99.8}} & 96.6 & 97.6 & 97.7 \\

Lingbot-VA~\cite{li2026causal}    
    & \textit{5B}    
    & \ding{51}    
    & \underline{\textit{92.90}} 
    & \underline{\textit{91.50}} 
    & 98.5 & 99.6 & 97.2 & \textbf{98.5} & 98.5 \\

Fast-WAM~\cite{yuan2026fast}    
    & \textit{6B}    
    & \ding{55}    
    & 91.88 
    & \textbf{91.78} 
    & 98.2 & \textbf{100.0} & 97.0 & 95.2 & 97.6 \\

WLA-0    
    & \textit{2B}    
    & \ding{55}    
    & \textbf{92.94} 
    & 90.02 
    & \underline{\textit{99.0}}
    & \textbf{100.0}
    & 97.8
    & 97.6
    & \underline{\textit{98.6}} \\

 \,\,\, $-\mathcal{L}_\mathrm{wm}$
    & \textit{2B}    
    & \ding{55}    
    & 90.98 
    & 89.34
    & 98.4 
    & 99.6 
    & 97.0
    & 96.4 
    & 97.9 \\

 \,\,\, $+$\textit{TTS}    
    & \textit{2B}    
    & \ding{55}    
    & --
    & -- 
    & \textbf{99.2} 
    & \textbf{100.0}
    & \textbf{98.4}
    & \underline{\textit{97.8}}
    & \textbf{98.9} \\

\bottomrule[0.5mm]
\end{tabular}
\end{table*} 

\textbf{RoboTwin 2.0.} 
RoboTwin 2.0~\cite{chen2025robotwin} is a challenging bimanual manipulation benchmark comprising 50 tasks requiring coordinated dual-arm control. 
Following prior multi-task training protocols~\cite{li2026causal, bi2025motus, yuan2026fast}, we train WLA-0 on a mixed demonstration dataset with 2,500 clean-scene trajectories and 25,000 strongly randomized trajectories for 100k steps with a global batch size of 256. 
Language-based subtask prediction is disabled due to the benchmark's short task horizons.
Discarding the World Expert at inference leaves $\sim$2B active parameters. 
To quantify the World Expert's impact, we evaluate an ablation ``$-\mathcal{L}_\mathrm{wm}$'' trained solely with action loss.
Table~\ref{tab:robotwin_and_libero} reports the average success rates over 100 trials per task.
WLA-0 achieves a 92.94\% success rate in clean environments, remains highly robust under domain randomization, and matches or outperforms prior pipelines while using significantly fewer parameters and no embodied pretraining.
Furthermore, the World Expert ablation confirms that future-state prediction effectively refines action generation.



\textbf{LIBERO.}
We train WLA-0 on all four LIBERO~\cite{liu2023libero} suites, i.e., Spatial, Object, Goal, and Long, each containing 10 tasks with 50 demonstrations per task. A single model is trained across all suites for 100k steps with a batch size of 256 and evaluated over 50 trials per task. In practice, we observe strong performance after only 30k training steps, highlighting the training efficiency of WLA-0.
As shown in Table~\ref{tab:robotwin_and_libero}, WLA-0 achieves 98.6\% average success, outperforming all WAM and VLA baselines. Given RoboTwin's high evaluation cost, we evaluate test-time scaling mainly on LIBERO. With test-time scaling using 6 candidates and an imagination horizon of 2, the average success rate further improves to 98.9\%.

\begin{table*}[t]
\caption{
\textbf{Comparison on RMBench.}
WLA-0 achieves the best performance on long-horizon, memory-dependent bimanual manipulation tasks.
\textbf{Bold} and \underline{\textit{Italics}} denote the best and second-best results, respectively.
$-\mathcal{L}_{\mathrm{lang}}$ denotes the variant without the language-based subtask prediction loss.
}
\renewcommand{\arraystretch}{1.2}
\setlength{\tabcolsep}{8pt} 
\centering
\small
\label{tab:rmbench}

\begin{tabular}{l|ccccc}
\toprule[0.5mm]
Method 
& \makecell{\textit{Battery Try}} 
& \makecell{\textit{Blocks Ranking Try}} 
& \textit{Cover Blocks} 
& \textit{Press Button} 
& \textit{\textbf{Average}} \\
\midrule



$\pi_{0.5}$    
    & 16\% 
    & 6\% 
    & 0\% 
    & 0\% 
    & 5.5\% \\

X-VLA    
    & 26\% 
    & 1\% 
    & 2\% 
    & 0\% 
    & 7.3\% \\

Mem-0    
    & 28\% 
    & 18\% 
    & \underline{\textit{68\%}} 
    & 0\% 
    & \underline{\textit{28.5\%}} \\

Fast-WAM    
    & 16\% 
    & \textbf{37\%} 
    & 0\% 
    & 0\% 
    & 13.3\% \\

WLA-0    
    & \textbf{45\%}
    & \underline{\textit{23\%}}
    & \textbf{84\%}
    & \textbf{74\%}
    & \textbf{56.5\%} \\

\,\,\,\,\, $-\mathcal{L}_{\mathrm{lang}}$
    & \underline{\textit{38\%}} 
    & 12\% 
    & 18\% 
    & \underline{\textit{1\%}} 
    & 17.3\% \\

\bottomrule[0.5mm]
\end{tabular}
\end{table*}

\textbf{RMBench.}
RMBench~\cite{chen2026rmbench} is a long-horizon, memory-dependent benchmark for bimanual manipulation. We evaluate WLA-0 on its $M(n)$ subset, whose four tasks require repeated exploration, trial-and-error recovery, long-term memory, and inference of the currently executable subtask from interaction history. These properties make RMBench well suited for assessing WLA-0's language-based reasoning and memory utilization. Following the RMBench protocol, we adopt a single-task training setup, training one model per task for 30k steps with a global batch size of 448. During inference, WLA-0 predicts textual subtasks conditioned on the instruction, initial frame, current observation, and accumulated subtask history; these predicted subtasks serve as memory traces that guide subsequent action generation.

We compare WLA-0 with representative baselines, including VLA baselines $\pi_{0.5}$~\cite{intelligence2025pi_} and X-VLA~\cite{zheng2025x}, the memory-based baseline Mem-0~\cite{chen2026rmbench}, and the WAM baseline Fast-WAM~\cite{yuan2026fast}. To assess the effect of subtask supervision, we also train an ablated variant, ``$-\mathcal{L}_{\mathrm{lang}}$'', which removes the subtask prediction loss. As shown in Table~\ref{tab:rmbench}, WLA-0 achieves the best average success rate of 56.5\%, substantially outperforming Fast-WAM and nearly doubling Mem-0. This result highlights the advantage of augmenting WAM-style action generation with language-level subtask reasoning and explicit memory tracking. The ablation result further confirms the importance of subtask supervision: removing the subtask prediction loss reduces the average success rate from 56.5\% to 17.25\%. These results show that semantic-level textual subtasks provide an effective interface for long-horizon progress tracking and action generation in memory-dependent manipulation. We provide a more detailed analysis and additional results for the simulation benchmarks in Appendix~\ref{app:simulation benchmarks}.

\subsection{Real-World Experiments}

\begin{figure}[t]
    \centering
    \includegraphics[width=1.0\linewidth]{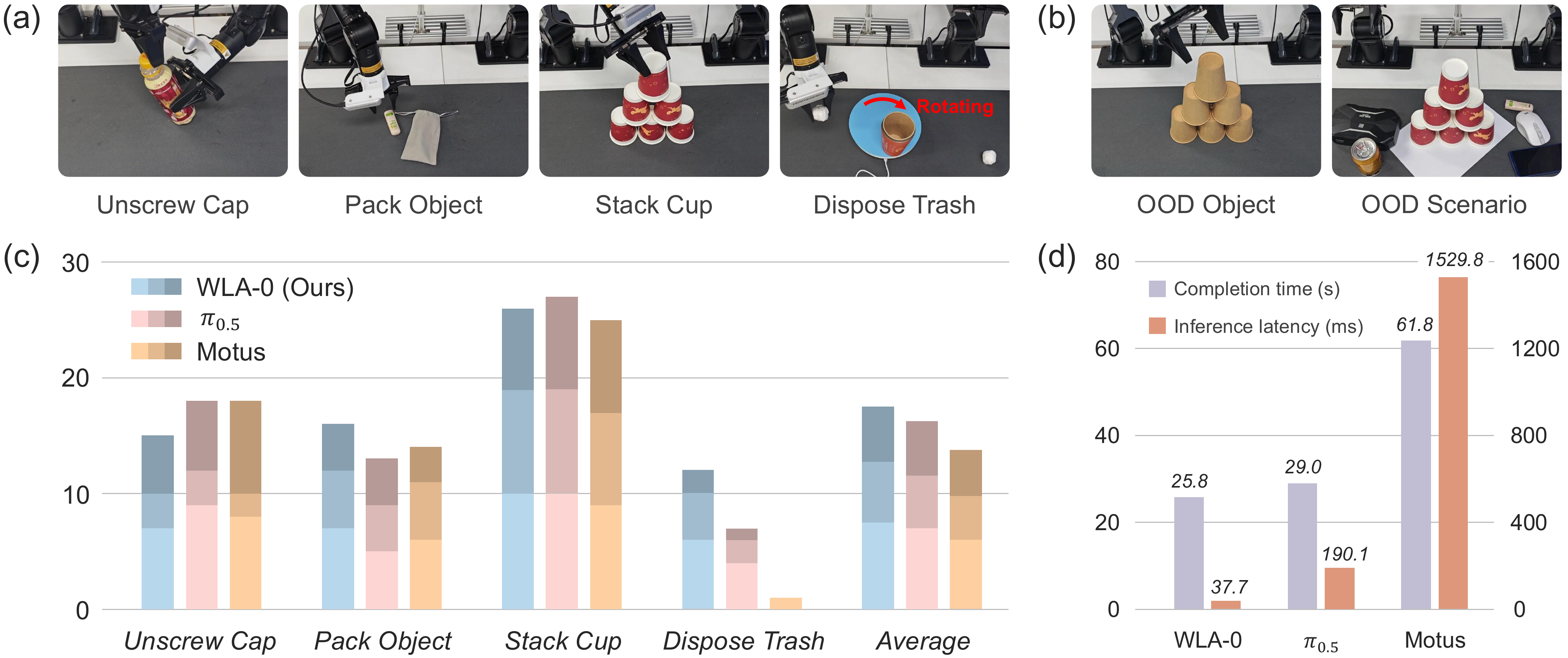}
    \caption{
    \textbf{Real-World Experiments.} (a) Illustrations of the four long-horizon tasks.
    (b) Examples of OOD Object and OOD Scenario tasks.
    (c) Success count comparison. Bar colors, from light to dark, denote the standard setup, OOD object, and OOD scenario, respectively.
    (d) Inference efficiency comparison. The left y-axis shows task completion time (s), and the right y-axis shows inference latency (ms).
    }
    \label{fig:real_world_experiments}
\end{figure}

\textbf{Task Setup.} We designed four long-horizon tasks to evaluate the real-world performance of WLA-0 (Fig.~\ref{fig:real_world_experiments} (a)).
\textit{Unscrew Cap} and \textit{Pack Object} evaluate fine-grained manipulation, while \textit{Stack Cup} and \textit{Dispose Trash} assess inference efficiency.
In \textit{Dispose Trash}, the robot must grasp scattered paper balls and place them into a trash bin rotating at varying speeds, requiring accurate trajectory prediction and low-latency control.
For each task, we collect 60 demonstrations using the AgilexRobotics Piper dual-arm platform. 
Each task is evaluated under three settings: (1) standard setup, (2) out-of-distribution (OOD) objects with novel shapes or colors (Fig.~\ref{fig:real_world_experiments} (b) \textit{left}), and (3) OOD scenarios with novel clutter or backgrounds (Fig.~\ref{fig:real_world_experiments} (b) \textit{right}).

\textbf{Success Count Comparison.} 
We compare WLA-0 (trained from scratch) with two pretrained baselines, $\pi_{0.5}$~\cite{intelligence2025pi_} and Motus~\cite{bi2025motus}. 
All models are trained for 50k steps with a global batch size of 256.  
Inference is performed on a single NVIDIA RTX 5090 GPU under a synchronous execution mode, ensuring the robot executes actions only after the inference cycle completes.
We report the average success count over 10 trials for each setting.
As shown in Fig.~\ref{fig:real_world_experiments} (c), WLA-0 performs comparably to the pretrained baselines,
and demonstrates robust generalization across OOD Object and Scenario settings despite lacking pretraining.
Notably, WLA-0 achieves the highest success rate on the dynamic \textit{Dispose Trash} task. 
This performance is attributed to its integration of historical context and future state prediction, which enables accurate environmental modeling. Combined with low inference latency, WLA-0 effectively adapts to real-time environmental changes. 
In contrast, Motus loses track of the rotating bin due to high inference latency, 
while $\pi_{0.5}$ misestimates the turntable velocity due to the absence of history conditioning, leading to task failure.

\textbf{Inference Efficiency Evaluation.}
To evaluate inference efficiency, we compare WLA-0, $\pi_{0.5}$, and Motus on the \textit{Stack Cup} task using two metrics: \textit{completion time} (s), measured from motion onset to task completion and averaged over 10 successful rollouts, and \textit{inference latency} (ms), averaged across all inference calls in those rollouts. 
All models use 10 flow-matching inference steps and an action chunk size of 32; to reduce GPU memory usage, Motus generates only one frame per inference call.
As shown in Figure~\ref{fig:real_world_experiments} (d), Motus exhibits the highest inference latency and the longest completion time, exceeding 60 seconds.
In contrast, WLA-0 achieves the lowest completion time and inference latency, consistently outperforming the VLA baseline $\pi_{0.5}$.
Notably, WLA-0 reduces inference latency by $\sim40\times$ relative to Motus, highlighting its suitability for real-time deployment, where high throughput and rapid reactivity are critical. See Appendix~\ref{app:Inference Efficiency Evaluation} for details.


\subsection{Learning New Tasks from Videos}
\begin{table*}[t]

\caption{
    \textbf{Comparison on five unseen RoboTwin 2.0 tasks.}
    We compare four settings: (1) \textit{Seen-Action}, trained only with action supervision from seen tasks; (2) \textit{Seen-Action}$+$\textit{Video}, which adds video supervision from seen tasks; (3) $+$\textit{Unseen Same-Emb. Video}, which further uses unseen-task videos from the same embodiment; and (4) $+$\textit{Unseen Cross-Emb. Video}, which uses unseen-task videos from the cross-embodiment robot.
    Each entry reports the success rates under \textit{Clean / Rand.} settings, respectively.
    \textbf{Bold} and \underline{\textit{Italics}} mark the best and second-best results.
}
\renewcommand{\arraystretch}{1.25}
\setlength{\tabcolsep}{6pt} 
\centering
\small
\label{tab:ablation}

\begin{tabular}{l|cccc}
\toprule[0.5mm]

Unseen Tasks                    
    & \makecell{\textit{Seen-Action} \\ \textit{(Baseline)}} 
    & \makecell{\textit{Seen-Action} \\ $+$\textit{Video}}
    & \makecell{$+$\textit{Unseen} \\ \textit{Same-Emb. Video}}
    & \makecell{$+$\textit{Unseen} \\ \textit{Cross-Emb. Video}}
    \\ 

\midrule

Beat Block Hammer     
    & 1 / 0         
    & 1 / 0          
    & \textbf{12} / \textbf{6}\,\,\,               
    & \underline{\textit{5}} / \underline{\textit{3}}
    \\

Move Playingcard Away 
    & \underline{\textit{2}} / 0         
    & 0 / \underline{\textit{3}}          
    & \textbf{42} / \textbf{28}              
    & 1 / 0                     
    \\

Pick Diverse Bottles  
    &  \,\,\,7 / 10        
    & 10 / 8\,\,\,         
    & \underline{\textit{32}} / \underline{\textit{29}}              
    & \textbf{39} / \textbf{35}                   
    \\

Place Object Basket   
    & 3 / 5         
    & 0 / 1          
    & \underline{\textit{30}} / \underline{\textit{34}}              
    & \textbf{45} / \textbf{47}                   
    \\
    
Stack Bowls Three     
    & 52 / 43       
    & 48 / 51        
    & \textbf{56} / \textbf{53}              
    & \underline{\textit{54}} / \underline{\textit{52}}                   
    \\

\textbf{Average}                 
    & 13.0 / 11.6   
    & 11.8 / 12.6    
    & \textbf{34.4} / \textbf{30.0}          
    & \underline{\textit{28.8}} / \underline{\textit{27.4}}               
    \\ 

\bottomrule[0.5mm]
\end{tabular}
\end{table*}


\begin{figure}[b]
    \centering
    \includegraphics[width=1.0\linewidth]{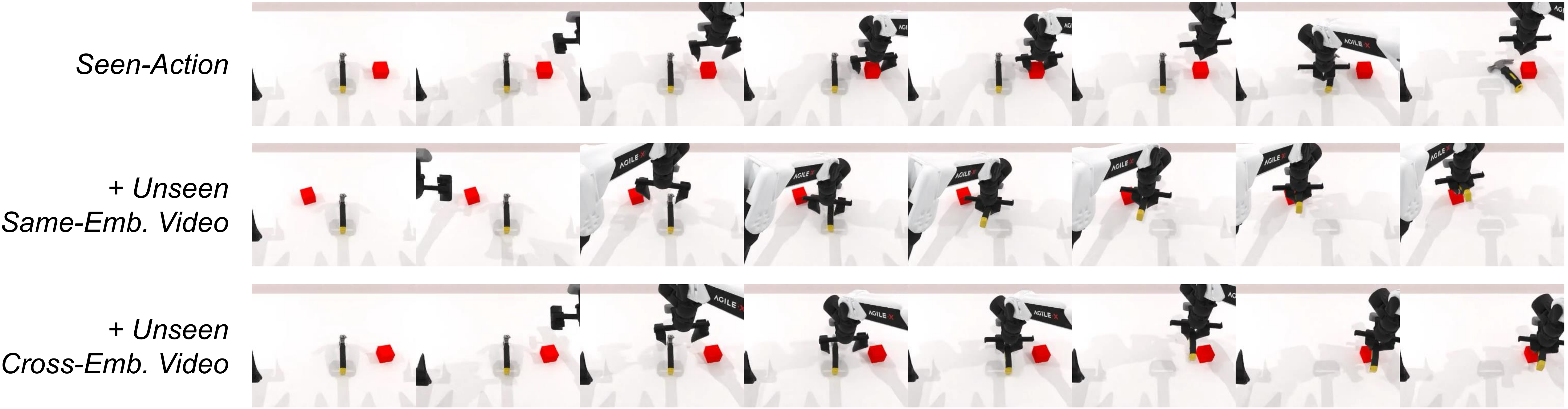}
    \caption{
    Visualization of the \textit{Beat Block Hammer} task execution under three settings: \textit{Seen-Action}, $+$\textit{Unseen Same-Emb. Video}, and $+$\textit{Unseen Cross-Emb. Video}.
    }
    \label{fig:learning_tasks_from_videos}
\end{figure}

The high cost of collecting action-annotated trajectories for specific robots bottlenecks dataset scalability. A scalable alternative instead learns novel skills from action-free videos of heterogeneous robots.
To evaluate whether WLA-0 exhibits this capability, we conduct experiments on the 50 tasks in RoboTwin2.0. We partition the tasks into 45 seen tasks and 5 unseen tasks, with each task containing 50 clean trajectories and 500 randomized trajectories. The unseen tasks cover five distinct action patterns, as summarized in Table~\ref{tab:ablation}.
We train models under four settings:
(1) \textit{Seen-Action} is the baseline, using only action supervision from seen tasks.
(2) \textit{Seen-Action}$+$\textit{Video} additionally uses video supervision from the seen tasks.
Building on this setting, (3) $+$\textit{Unseen Same-Emb. Video} further incorporates video supervision from the five unseen tasks under the same embodiment, Aloha-AgileX. 
In contrast, (4) $+$\textit{Unseen Cross-Emb. Video} uses video supervision from unseen tasks under a cross-embodiment robot, ARX-X5. 

As reported in Table~\ref{tab:ablation}, the \textit{Seen-Action} baseline and \textit{Seen-Action}$+$\textit{Video} struggle significantly, failing almost entirely on tasks like \textit{Beat Block Hammer} and \textit{Move Playingcard Away}.
In contrast, $+$\textit{Unseen Same-Emb. Video} nearly triples the baseline success rate to achieve the best overall results.
Remarkably, the model acquires the novel ``beat'' action solely from video observations. In Beat Block Hammer, for instance, it correctly grasps the hammer and attempts to strike the target, whereas the baseline erroneously attempts to grasp the block directly, as shown in Fig.~\ref{fig:learning_tasks_from_videos}.
Furthermore, $+$\textit{Unseen Cross-Emb. Video} maintains a highly competitive success rate, highlighting WLA-0’s ability to align visual observations with actionable control knowledge, demonstrating robust cross-embodiment transfer, cross-modal learning, and task generalization.
Demos and additional studies on learning from human egocentric videos are detailed in Appendix~\ref{app:human video}.

\section{Conclusion}
\label{sec:conclusion}
We introduced WLA, a unified embodied framework that integrates world modeling, language reasoning, and action synthesis. 
Using an autoregressive language backbone with the World Expert and Action Expert, WLA models semantic-level textual subtasks and fine-grained physical dynamics, enabling effective long-horizon reasoning and real-time robot control. Extensive experiments show that WLA-0 achieves strong multi-task performance, state-of-the-art results on memory-dependent manipulation tasks, and favorable inference efficiency. Moreover, its ability to learn new tasks from action-free videos suggests a promising direction for scalable cross-embodiment robot learning.

\section{Limitations}
\label{sec:limitations}

Despite its promising results, WLA has several limitations. First, the real-world experiments are currently limited to a small set of bimanual tasks on a single robot platform; broader evaluations across diverse embodiments and task domains are needed to further establish its generality. In addition, our video-based task learning experiments rely on simulated robot videos for supervision.

\setlength{\bibsep}{5pt}
\bibliography{reference}
\bibliographystyle{plainnat}

\newpage
\appendix

\section{Acceleration Techniques}
\label{app:acceleration techniques}
WLA-0’s inference latency is dominated by Python dispatch overhead and the cost of launching many small CUDA kernels, especially in the iterative DiT denoising loop. We address these bottlenecks with three complementary optimizations, reducing latency from $\sim$116 ms to under 40 ms:

\textbf{CUDA Graph Capture.} We replace eager execution with CUDA Graph replay. Specifically, we capture the forward pass once using fixed-address GPU buffers and replay the captured graph for subsequent inference calls. This removes per-step Python dispatch and substantially reduces kernel-launch overhead, which is especially beneficial for the multi-step DiT head.

\textbf{Operator Fusion.} We implement custom Triton kernels to fuse frequently adjacent operations, reducing both launch overhead and intermediate memory traffic. In the VLM, we fuse RMSNorm~\cite{zhang2019root}, QKV projection with per-head RMSNorm and RoPE~\cite{su2024roformer}, the SwiGLU~\cite{shazeer2020glu} branch, and decoder-layer computations. In the DiT, we fuse AdaLayerNorm~\cite{peebles2023scalable}, merged-QKV self-attention, cross-attention with cached K/V, position-embedding addition, and the GELU~\cite{hendrycks2016gaussian} feed-forward block.

\textbf{Precomputation and Caching.} We precompute quantities that remain invariant across inference calls or denoising steps. In the VLM, these include token embeddings, causal masks, RoPE sine/cosine tables, and image-placeholder indices. In the DiT, they include sinusoidal action-time encodings, timestep-MLP outputs, and AdaLN scale/shift parameters. We also cache cross-attention K/V tensors and reuse them across all denoising steps for a given prediction.



\section{Simulation Benchmarks}
\label{app:simulation benchmarks}

\subsection{RoboTwin 2.0}
In Table~\ref{tab:robotwin_and_libero}, LingBot-VA~\cite{li2026causal} is evaluated under the \textit{seen instructions} setting, whereas the other methods use the \textit{unseen instructions} setting. 
Per-task results are reported in Table~\ref{tab:robotwin_per_task}. 
For WLA-0, we use 32 flow-matching inference steps, as preliminary experiments showed that fewer steps can induce robotic-arm jitter. 
Actions are represented by the absolute end-effector position.

\subsection{LIBERO} 
\begin{table}[b]

\caption{
\textbf{Comparison of single-frame and multi-frame prediction on LIBERO.}
Multi-frame prediction yields a substantially lower success rate than single-frame prediction.
}
\vspace{5pt}
\label{tab:app_libero}
\centering
\renewcommand{\arraystretch}{1.25}
\setlength{\tabcolsep}{6pt} 

\begin{tabular}{lccccc}
\toprule[0.5mm]
& \textit{Spatial} & \textit{Object} & \textit{Goal} & \textit{Long} & \textbf{\textit{Avg.}} 
\\ 
\midrule
single-frame 
& 98.8    & 100.0  & 97.4 & 96.6 & 98.2 \\
multi-frame  
& 95.4    & 96.6   & 93.2 & 91.4 & 94.2 \\ 
\bottomrule[0.5mm]
\end{tabular}
\end{table}

We conducted preliminary experiments to examine how the World Expert’s prediction target (single-frame or multi-frame) affects action learning.
For $n=32$, the single-frame model predicts only $\mathbf{o}_{t+32}$, while the multi-frame model jointly predicts $\mathbf{o}_{t+8}$, $\mathbf{o}_{t+16}$, $\mathbf{o}_{t+24}$, and $\mathbf{o}_{t+32}$. 
Both models were trained for 30k steps with a global batch size of 256. 
As shown in Table~\ref{tab:app_libero}, multi-frame prediction achieves a substantially lower success rate than single-frame prediction, suggesting that overly dense visual supervision may slow convergence and interfere with action learning.

\subsection{RMBench}

\begin{figure*}[t]
    \centering

    \includegraphics[width=0.95\textwidth]{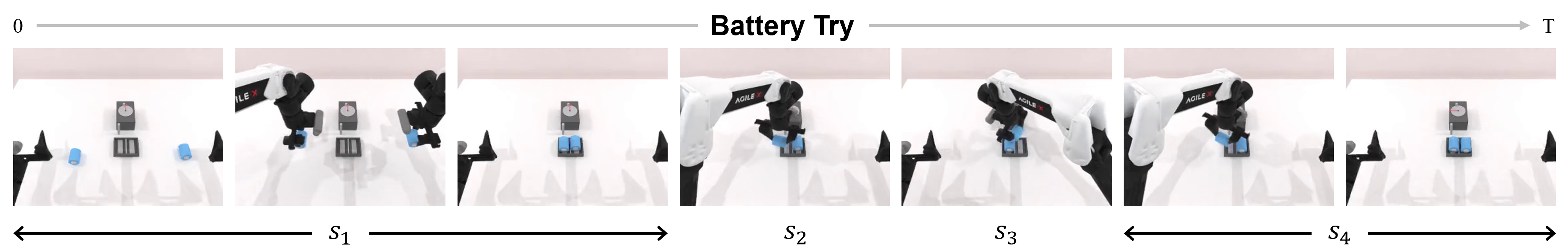}

    \begin{tcolorbox}[
        title=Battery Try,
        fonttitle=\bfseries,
        colback={rgb,255:red,245;green,248;blue,250},
        colframe={rgb,255:red,144;green,162;blue,179},
        boxrule=1pt,
        width=0.95\textwidth
    ]
    \begin{lstlisting}[style=promptstyle]
(*@\lstheading{Task instruction:}@*)
There are two batteries and a battery slot on the table. Combining the two batteries in different orientations causes the dashboard needle to rotate.

(*@\lstheading{Subtask decomposition:}@*)
(*@\subtasklabel{1}@*) Use dual arm to pick up the batteries and place them into the battery slots in the positive direction.
(*@\subtasklabel{2}@*) Pick up the left battery and place it into the battery slot in the negative direction.
(*@\subtasklabel{3}@*) Pick up the right battery and place it into the battery slot in the negative direction.
(*@\subtasklabel{4}@*) Pick up the left battery and place it into the battery slot in the positive direction.
    \end{lstlisting}
    \end{tcolorbox}

    \caption{
    Illustration of the Battery Try task in RMBench.
    Top: representative frames along the task trajectory.
    Bottom: task instruction and subtask decomposition.
    }
    \label{fig:rmbench_battery_try_combined}
\end{figure*}
\begin{figure*}[t]
    \centering

    \includegraphics[width=0.95\textwidth]{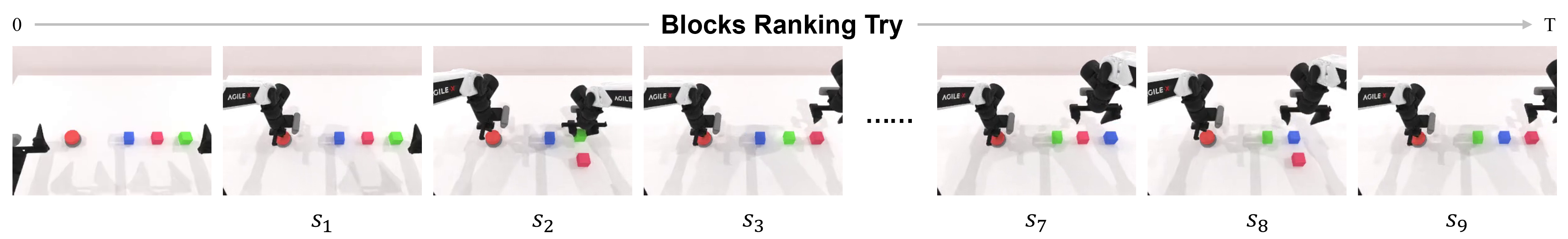}

    \begin{tcolorbox}[
        title=Blocks Ranking Try,
        fonttitle=\bfseries,
        colback={rgb,255:red,245;green,248;blue,250},
        colframe={rgb,255:red,144;green,162;blue,179},
        boxrule=1pt,
        width=0.95\textwidth
    ]
    \begin{lstlisting}[style=promptstyle]
(*@\lstheading{Task instruction:}@*)
There is a button and three colored cubes arranged in a random row on the table. Each time the cubes are rearranged, the arm presses the button until the arrangement is successful.

(*@\lstheading{Subtask decomposition:}@*)
(*@\subtasklabel{1}@*) The colors of the blocks from left to right are blue, pink, and green; press the button.
(*@\subtasklabel{2}@*) Swap the middle block and the right block.
(*@\subtasklabel{3}@*) The colors of the blocks from left to right are blue, green, and pink; press the button.
(*@\subtasklabel{4}@*) Swap the left block and the right block.
(*@\subtasklabel{5}@*) The colors of the blocks from left to right are pink, green, and blue; press the button.
(*@\subtasklabel{6}@*) Swap the left block and the middle block.
(*@\subtasklabel{7}@*) The colors of the blocks from left to right are green, pink, and blue; press the button.
(*@\subtasklabel{8}@*) Swap the middle block and the right block.
(*@\subtasklabel{9}@*) The colors of the blocks from left to right are green, blue, and pink; press the button.
    \end{lstlisting}
    \end{tcolorbox}

    \caption{
    Illustration of the Blocks Ranking Try task in RMBench.
    Top: representative frames along the task trajectory.
    Bottom: task instruction and subtask decomposition.
    }
    \label{fig:rmbench_blocks_ranking_try_combined}
\end{figure*}
\begin{figure*}[t]
    \centering

    \includegraphics[width=0.95\textwidth]{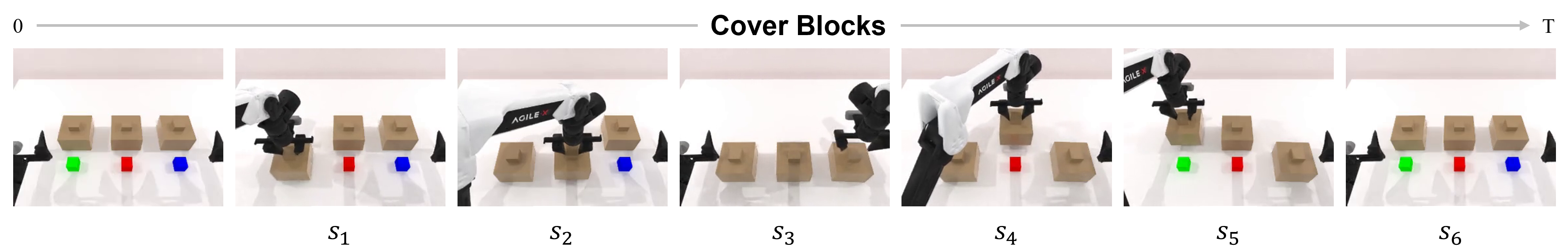}

    \begin{tcolorbox}[
        title=Cover Blocks,
        fonttitle=\bfseries,
        colback={rgb,255:red,245;green,248;blue,250},
        colframe={rgb,255:red,144;green,162;blue,179},
        boxrule=1pt,
        width=0.95\textwidth
    ]
    \begin{lstlisting}[style=promptstyle]
(*@\lstheading{Task instruction:}@*)
On the table, red, green, and blue blocks are arranged randomly along with three lids. From the current viewpoint, cover the blocks from left to right using the lids, and then uncover them again in the sequence red, green, and blue.

(*@\lstheading{Subtask decomposition:}@*)
(*@\subtasklabel{1}@*) Cover the left block with the left cover.
(*@\subtasklabel{2}@*) Cover the middle block with the middle cover.
(*@\subtasklabel{3}@*) Cover the right block with the right cover.
(*@\subtasklabel{4}@*) Open the middle cover to uncover the blocks in the order of red, green, and blue.
(*@\subtasklabel{5}@*) Open the left cover to uncover the blocks in the order of red, green, and blue.
(*@\subtasklabel{6}@*) Open the right cover to uncover the blocks in the order of red, green, and blue.
    \end{lstlisting}
    \end{tcolorbox}

    \caption{
    Illustration of the Cover Blocks task in RMBench.
    Top: representative frames along the task trajectory.
    Bottom: task instruction and subtask decomposition.
    }
    \label{fig:rmbench_cover_blocks_combined}
\end{figure*}
\begin{figure*}[t]
    \centering

    \includegraphics[width=0.95\textwidth]{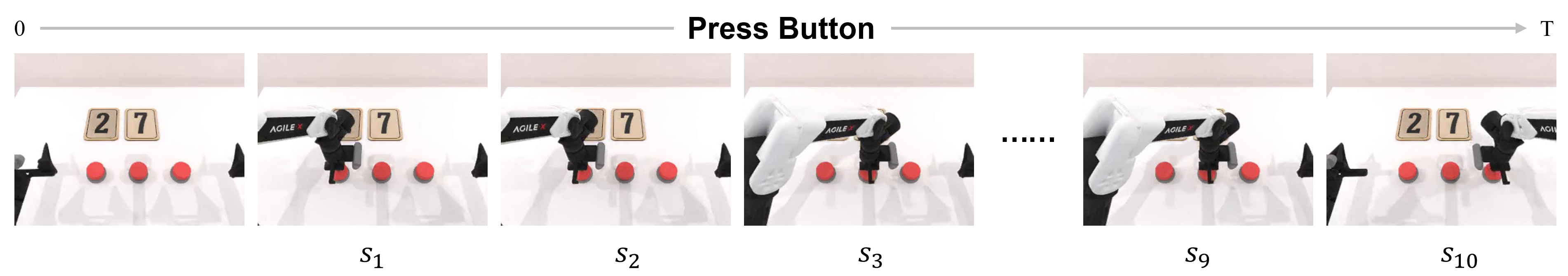}

    \begin{tcolorbox}[
        title=Press Button,
        fonttitle=\bfseries,
        colback={rgb,255:red,245;green,248;blue,250},
        colframe={rgb,255:red,144;green,162;blue,179},
        boxrule=1pt,
        width=0.95\textwidth
    ]
    \begin{lstlisting}[style=promptstyle]
(*@\lstheading{Task instruction:}@*)
Observe the two numbers on the table. Press the left button the number of times corresponding to the number on the left, and press the middle button the number of times corresponding to the number on the right. Then press the right button once to confirm.

(*@\lstheading{Subtask decomposition:}@*)
(*@\subtasklabel{1}@*) Press the left button for the first time.
(*@\subtasklabel{2}@*) Press the left button for the second time.
(*@\subtasklabel{3}@*) Press the middle button for the first time.
(*@\subtasklabel{4}@*) Press the middle button for the second time.
(*@\subtasklabel{5}@*) Press the middle button for the third time.
(*@\subtasklabel{6}@*) Press the middle button for the fourth time.
(*@\subtasklabel{7}@*) Press the middle button for the fifth time.
(*@\subtasklabel{8}@*) Press the middle button for the sixth time.
(*@\subtasklabel{9}@*) Press the middle button for the seventh time.
(*@\subtasklabel{10}@*) Press the confirm button.
    \end{lstlisting}
    \end{tcolorbox}

    \caption{
    Illustration of the Press Button task in RMBench.
    Top: representative frames along the task trajectory.
    Bottom: task instruction and subtask decomposition.
    }
    \label{fig:rmbench_press_button_combined}
\end{figure*}

Figures~\ref{fig:rmbench_battery_try_combined}--\ref{fig:rmbench_press_button_combined} provide additional RMBench task illustrations, respectively showing the trajectories and language-level subtask decompositions of Battery Try, Blocks Ranking Try, Cover Blocks, and Press Button.
As shown in Table~\ref{tab:rmbench}, WLA-0 achieves the best average success rate on RMBench. $\pi_{0.5}$, X-VLA, and Fast-WAM mainly generate actions from visual observations and instructions, but lack explicit memory traces and language-level progress planning. Therefore, they struggle to infer the current executable subtask when the next action depends on previous trials. Compared with Mem-0, the advantage of WLA-0 mainly comes from tighter synchronization between progress tracking and action generation. Mem-0 relies on a separate Subtask End Classifier to detect visually subtle subtask transitions; if the classifier makes an incorrect transition decision, the memory can be updated too early or too late, leading to incorrect subtask switching and affecting subsequent action generation. In contrast, WLA-0 repeatedly infers the current executable subtask before action generation, leading to more stable progress tracking.

\section{Real-World Experiments}
\label{app:Inference Efficiency Evaluation}
\begin{table}[t]
\centering
\caption{
Evaluation under standard and OOD settings on four real-world tasks. 
}
\vspace{5pt}
\label{tab:real_world_details}
\setlength{\tabcolsep}{10pt}
\renewcommand{\arraystretch}{1.05}
\begin{tabular}{llccc}
\toprule[0.5mm]
\textbf{Task} & \textbf{Setting} & \textbf{WLA-0} & \boldmath$\pi_{0.5}$ & \textbf{Motus} \\
\midrule
\multirow{3}{*}{\textit{Unscrew Cap}}
& {\small Standard} & 7 & 9 & 8 \\
& {\small\textit{OOD Object}} & 3 & 3 & 2 \\
& {\small\textit{OOD Scenario}} & 6 & 6 & 5 \\
\midrule

\multirow{3}{*}{\textit{Pack Object}}
& {\small Standard} & 7 & 5 & 6 \\
& {\small\textit{OOD Object}} & 5 & 4 & 5 \\
& {\small\textit{OOD Scenario}} & 4 & 4 & 3 \\
\midrule

\multirow{3}{*}{\textit{Stack Cup}}
& {\small Standard} & 10 & 10 & 9 \\
& {\small\textit{OOD Object}} & 9 & 9 & 8 \\
& {\small\textit{OOD Scenario}} & 7 & 8 & 8 \\
\midrule

\multirow{3}{*}{\textit{Dispose Trash}}
& {\small Standard} & 6 & 4 & 1 \\
& {\small\textit{OOD Object}} & 4 & 2 & 0 \\
& {\small\textit{OOD Scenario}} & 2 & 1 & 0 \\
\midrule

\multirow{3}{*}{\textbf{Avg.}}
& {\small Standard} & 7.5 & 7 & 6 \\
& {\small\textit{OOD Object}} & 5.25 & 4.5 & 3.75 \\
& {\small\textit{OOD Scenario}} & 4.75 & 4.75 & 4 \\
\bottomrule[0.5mm]
\end{tabular}
\end{table}

\begin{figure}[t]
    \centering
    \includegraphics[width=1.0\linewidth]{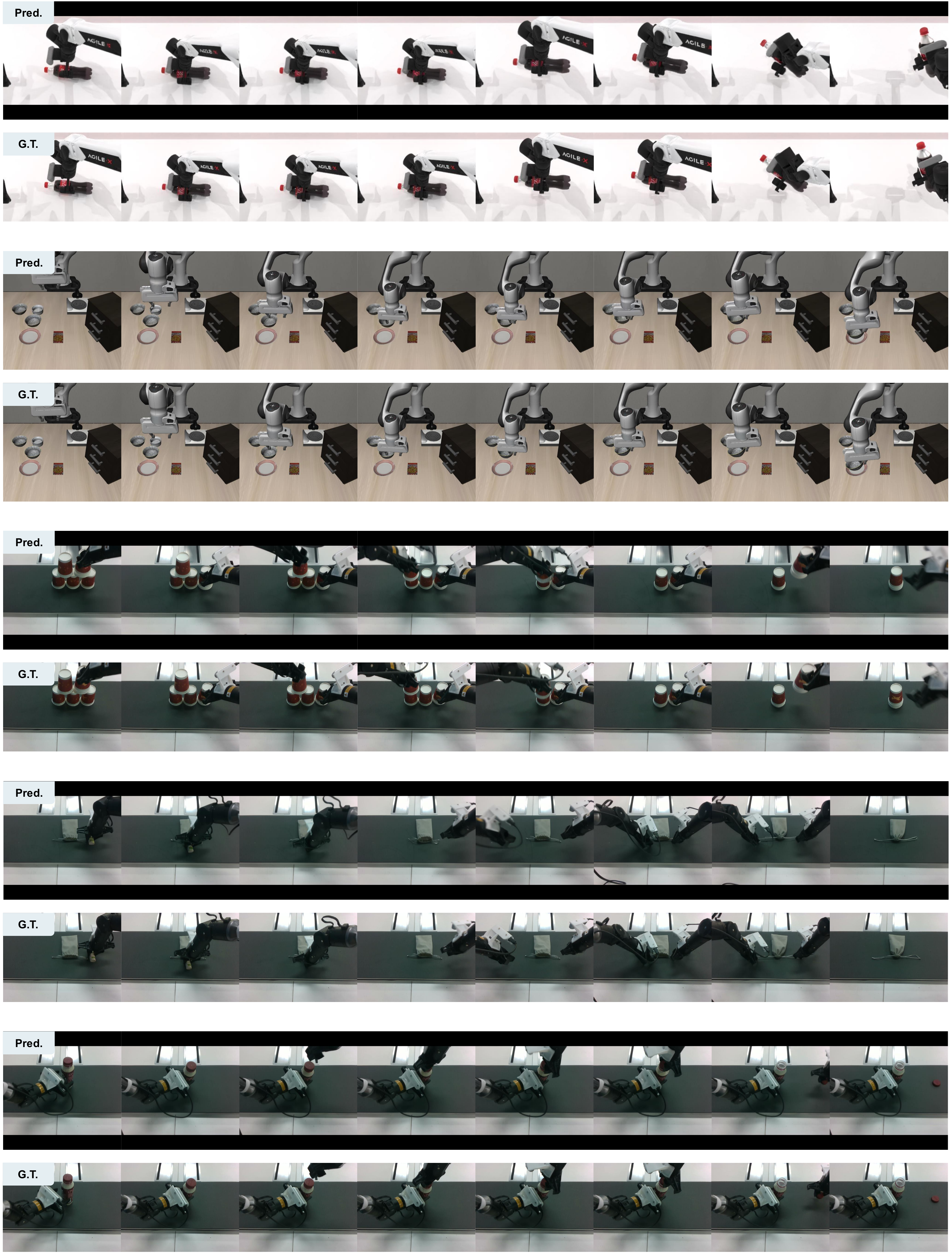}
    \caption{
    Visualization of the predicted (Pred.) and ground-truth (G.T.) images during inference.
    }
    \label{fig:appendix_world_expert_visual}
\end{figure}

For real-world robot experiments, actions are represented by the robot arm’s joint angles, and flow-matching inference is performed with 10 steps. The full results are reported in Table~\ref{tab:real_world_details}.
We further visualize the images generated by the World Expert during inference in Figure~\ref{fig:appendix_world_expert_visual}.

\section{Learning New Tasks from Videos}
\label{app:human video}



\begin{table*}[t]

\caption{
    \textbf{Comparison on five unseen RoboTwin 2.0 tasks.}
    We compare two settings: (1) \textit{Seen-Action} and (2) $+$\textit{Unseen Human-Ego. Video}, which uses unseen-task human egocentric videos.
    Each entry reports the success rates under \textit{Clean / Rand.} settings, respectively.
}
\renewcommand{\arraystretch}{1.25}
\setlength{\tabcolsep}{6pt} 
\centering
\label{tab:human_videos}

\begin{tabular}{l|cc}
\toprule[0.5mm]

Unseen Tasks                    
    & \makecell{\textit{Seen-Action} \\ \textit{(Baseline)}} 
    & \makecell{$+$\textit{Unseen} \\ \textit{Human-Ego. Video}}
    \\ 

\midrule

Beat Block Hammer     
    & 1 / 0         
    & 0 / 0          
    \\

Move Playingcard Away 
    & 2 / 0         
    & 0 / 3          
    \\

Pick Diverse Bottles  
    & 7 / 10        
    & 1 / 1         
    \\

Place Object Basket   
    & 3 / 5         
    & 17 / 14          
    \\
    
Stack Bowls Three     
    & 52 / 43       
    & 21 / 21        
    \\

\textbf{Average}                 
    & 13.0 / 11.6   
    & 7.8 / 7.8    
    \\ 

\bottomrule[0.5mm]
\end{tabular}
\end{table*}

\begin{figure}[t]
    \centering
    \includegraphics[width=1.0\linewidth]{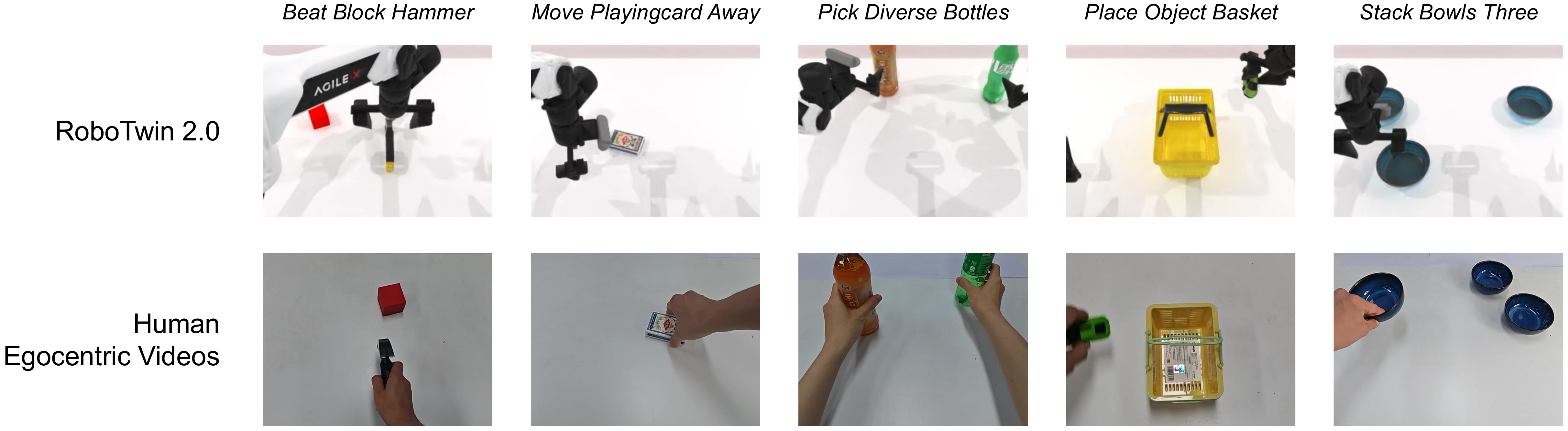}
    \caption{
    Unseen Tasks in RoboTwin 2.0 and corresponding Human Egocentric Videos.
    }
    \vspace{1em}
    \label{fig:human_videos}
\end{figure}

For training in the $+$\textit{Unseen Same-Emb. Video} and $+$\textit{Unseen Cross-Emb. Video} settings, we set the loss weight for videos from unseen tasks to 0.1 and sample data from seen tasks and unseen-task videos at a 1:1 ratio in each forward pass. All models are trained for 50k steps with a global batch size of 256.
We further investigated whether WLA-0 can learn unseen tasks from human egocentric videos. We collected a set of real-world props that resemble the objects in the RoboTwin 2.0 simulation environment and recorded 100 human egocentric manipulation videos for each of the five unseen tasks, as shown in Figure~\ref{fig:human_videos}. We then mixed these videos with data from the seen tasks for training. The results are reported in Table~\ref{tab:human_videos}.
However, adding human egocentric videos did not enable the model to learn the new tasks. We conjecture that this failure is primarily due to the domain gap between real-world human videos and the simulation environment. In future work, we plan to better align human-video data with robot data and further validate this hypothesis.

\begin{table*}[t]
\centering
\caption{
    \textbf{Per-task success rates on RoboTwin 2.0. }
    \textbf{Bold} denotes the best results.
}
\label{tab:robotwin_per_task}
\scriptsize
\setlength{\tabcolsep}{2.5pt}
\renewcommand{\arraystretch}{0.95}
\resizebox{\textwidth}{!}{%
\begin{tabular}{l*{12}{c}}
\toprule[0.5mm]
\textbf{Task}
& \multicolumn{2}{c}{\textbf{WLA-0}}
& \multicolumn{2}{c}{\textbf{\makecell{$-\mathcal{L}_\mathrm{wm}$}}}
& \multicolumn{2}{c}{\boldmath$\pi_{0.5}$}
& \multicolumn{2}{c}{\textbf{Motus}}
& \multicolumn{2}{c}{\textbf{LingBot-VA}}
& \multicolumn{2}{c}{\textbf{Fast-WAM}} \\[2pt]
& \multicolumn{1}{c}{\scriptsize{Clean}} & \multicolumn{1}{c}{\scriptsize{Rand.}}
& \multicolumn{1}{c}{\scriptsize{Clean}} & \multicolumn{1}{c}{\scriptsize{Rand.}}
& \multicolumn{1}{c}{\scriptsize{Clean}} & \multicolumn{1}{c}{\scriptsize{Rand.}}
& \multicolumn{1}{c}{\scriptsize{Clean}} & \multicolumn{1}{c}{\scriptsize{Rand.}}
& \multicolumn{1}{c}{\scriptsize{Clean}} & \multicolumn{1}{c}{\scriptsize{Rand.}}
& \multicolumn{1}{c}{\scriptsize{Clean}} & \multicolumn{1}{c}{\scriptsize{Rand.}} \\
\midrule
\textit{Adjust Bottle} & \textbf{100} & \textbf{100} & \textbf{100} & \textbf{100} & \textbf{100} & 99 & 89 & 93 & 90 & 94 & \textbf{100} & \textbf{100} \\
\textit{Beat Block Hammer} & 95 & 87 & 93 & 88 & 96 & 93 & 95 & 88 & 96 & \textbf{98} & \textbf{99} & 97 \\
\textit{Blocks Ranking RGB} & 98 & 98 & 95 & 94 & 92 & 85 & 99 & 97 & 99 & 98 & \textbf{100} & \textbf{100} \\
\textit{Blocks Ranking Size} & 93 & 85 & 82 & 83 & 49 & 26 & 75 & 63 & \textbf{94} & 96 & \textbf{94} & \textbf{98} \\
\textit{Click Alarmclock} & 99 & \textbf{100} & 99 & 98 & 98 & 89 & \textbf{100} & \textbf{100} & 99 & \textbf{100} & \textbf{100} & \textbf{100} \\
\textit{Click Bell} & \textbf{100} & \textbf{100} & \textbf{100} & \textbf{100} & 99 & 66 & \textbf{100} & \textbf{100} & \textbf{100} & \textbf{100} & \textbf{100} & \textbf{100} \\
\textit{Dump Bin Bigbin} & 90 & 94 & 90 & 94 & 92 & \textbf{97} & 95 & 91 & 89 & 96 & \textbf{97} & 96 \\
\textit{Grab Roller} & \textbf{100} & \textbf{100} & \textbf{100} & \textbf{100} & \textbf{100} & \textbf{100} & \textbf{100} & \textbf{100} & \textbf{100} & \textbf{100} & \textbf{100} & \textbf{100} \\
\textit{Handover Block} & 96 & \textbf{87} & \textbf{100} & 80 & 66 & 57 & 86 & 73 & 99 & 78 & 95 & 81 \\
\textit{Handover Mic} & 92 & 93 & 91 & 94 & 98 & 97 & 78 & 63 & 94 & 96 & \textbf{99} & \textbf{100} \\
\textit{Hanging Mug} & \textbf{69} & 47 & 46 & 44 & 18 & 17 & 38 & 38 & 40 & 28 & 58 & \textbf{62} \\
\textit{Lift Pot} & \textbf{100} & \textbf{100} & 98 & \textbf{100} & 96 & 85 & 96 & 99 & \textbf{100} & 99 & \textbf{100} & \textbf{100} \\
\textit{Move Can Pot} & \textbf{98} & 99 & 97 & \textbf{100} & 51 & 55 & 34 & 74 & 94 & 97 & 90 & 88 \\
\textit{Move Pillbottle Pad} & \textbf{100} & 97 & 98 & 96 & 84 & 61 & 93 & 96 & 99 & \textbf{99} & \textbf{100} & \textbf{99} \\
\textit{Move Playingcard Away} & 99 & \textbf{100} & 99 & 99 & 96 & 84 & \textbf{100} & 96 & \textbf{100} & 99 & \textbf{100} & \textbf{100} \\
\textit{Move Stapler Pad} & \textbf{92} & 75 & 88 & 81 & 56 & 42 & 83 & \textbf{85} & 91 & 79 & 77 & 64 \\
\textit{Open Laptop} & \textbf{99} & \textbf{100} & 96 & 98 & 90 & 96 & 95 & 91 & 92 & 94 & 98 & \textbf{100} \\
\textit{Open Microwave} & \textbf{97} & \textbf{92} & 93 & \textbf{92} & 34 & 77 & 95 & 91 & 82 & 86 & 62 & 45 \\
\textit{Pick Diverse Bottles} & 95 & 79 & \textbf{96} & 86 & 81 & 71 & 90 & \textbf{91} & 89 & 82 & 80 & 85 \\
\textit{Pick Dual Bottles} & \textbf{100} & 83 & \textbf{100} & 95 & 93 & 63 & 96 & 90 & \textbf{100} & \textbf{99} & \textbf{100} & 96 \\
\textit{Place A2B Left} & 77 & 76 & 85 & 82 & 87 & 82 & 88 & 79 & \textbf{97} & \textbf{93} & 95 & \textbf{93} \\
\textit{Place A2B Right} & 75 & 75 & 72 & 75 & 87 & 84 & 91 & 87 & \textbf{97} & 95 & 93 & \textbf{99} \\
\textit{Place Bread Basket} & 91 & 91 & 92 & 91 & 77 & 64 & 91 & 94 & \textbf{97} & \textbf{95} & 91 & 93 \\
\textit{Place Bread Skillet} & 94 & 85 & 94 & 83 & 85 & 66 & 86 & 83 & \textbf{95} & 90 & 90 & \textbf{93} \\
\textit{Place Burger Fries} & 95 & 98 & 95 & 95 & 94 & 87 & \textbf{98} & 98 & 97 & 95 & 96 & \textbf{99} \\
\textit{Place Can Basket} & \textbf{87} & 78 & 85 & 80 & 62 & 62 & 81 & 76 & 81 & \textbf{84} & 71 & 69 \\
\textit{Place Cans Plasticbox} & \textbf{100} & 98 & 98 & 95 & 94 & 84 & 98 & 94 & \textbf{100} & \textbf{99} & 99 & 96 \\
\textit{Place Container Plate} & 99 & 99 & \textbf{100} & 98 & 99 & 95 & 98 & 99 & 99 & 97 & 96 & \textbf{100} \\
\textit{Place Dual Shoes} & \textbf{94} & 92 & 89 & \textbf{94} & 75 & 75 & 93 & 87 & \textbf{94} & 89 & \textbf{94} & 88 \\
\textit{Place Empty Cup} & 99 & \textbf{100} & \textbf{100} & \textbf{100} & \textbf{100} & 99 & 99 & 98 & \textbf{100} & \textbf{100} & \textbf{100} & \textbf{100} \\
\textit{Place Fan} & 94 & 94 & 95 & 90 & 87 & 85 & 91 & 87 & \textbf{99} & 93 & 96 & \textbf{96} \\
\textit{Place Mouse Pad} & 89 & 88 & 75 & 76 & 60 & 39 & 66 & 68 & \textbf{93} & \textbf{96} & 83 & 89 \\
\textit{Place Object Basket} & 82 & 84 & 85 & 84 & 80 & 76 & 81 & 87 & \textbf{91} & \textbf{88} & 89 & \textbf{88} \\
\textit{Place Object Scale} & \textbf{99} & 96 & 90 & 92 & 86 & 80 & 88 & 85 & 96 & 95 & 90 & \textbf{97} \\
\textit{Place Object Stand} & 99 & 92 & \textbf{100} & 91 & 91 & 85 & 98 & \textbf{97} & 99 & 96 & 90 & 94 \\
\textit{Place Phone Stand} & 95 & 98 & 91 & 97 & 81 & 81 & 87 & 86 & \textbf{97} & 97 & \textbf{97} & \textbf{99} \\
\textit{Place Shoe} & \textbf{100} & \textbf{99} & 97 & \textbf{99} & 92 & 93 & 99 & 97 & 98 & 98 & 96 & \textbf{99} \\
\textit{Press Stapler} & \textbf{99} & 97 & 83 & 85 & 87 & 83 & 93 & \textbf{98} & 85 & 82 & 90 & 97 \\
\textit{Put Bottles Dustbin} & 89 & 85 & 90 & 90 & 84 & 79 & 81 & 79 & 87 & \textbf{91} & \textbf{95} & 90 \\
\textit{Put Object Cabinet} & 82 & 84 & 80 & 79 & 80 & 79 & 88 & 71 & 85 & 87 & \textbf{94} & \textbf{89} \\
\textit{Rotate QRcode} & 91 & 91 & 93 & \textbf{94} & 89 & 87 & 89 & 73 & \textbf{96} & 91 & 93 & 89 \\
\textit{Scan Object} & \textbf{96} & \textbf{95} & 92 & 91 & 72 & 65 & 67 & 66 & \textbf{96} & 91 & 89 & 92 \\
\textit{Shake Bottle} & 99 & \textbf{100} & \textbf{100} & 98 & 99 & 97 & \textbf{100} & 97 & \textbf{100} & 97 & \textbf{100} & \textbf{100} \\
\textit{Shake Bottle Horizontally} & 99 & \textbf{100} & 98 & 97 & 99 & 99 & \textbf{100} & 98 & \textbf{100} & 99 & \textbf{100} & \textbf{100} \\
\textit{Stack Blocks Three} & 95 & 91 & 95 & 94 & 91 & 76 & 91 & 95 & \textbf{99} & \textbf{98} & 95 & 97 \\
\textit{Stack Blocks Two} & \textbf{100} & \textbf{100} & \textbf{100} & \textbf{100} & 97 & \textbf{100} & \textbf{100} & 98 & \textbf{100} & 98 & \textbf{100} & \textbf{100} \\
\textit{Stack Bowls Three} & \textbf{86} & 84 & 77 & 75 & 77 & 71 & 79 & \textbf{87} & \textbf{86} & 83 & 80 & 81 \\
\textit{Stack Bowls Two} & 96 & \textbf{99} & 94 & 97 & 95 & 96 & \textbf{98} & 98 & 94 & 98 & 92 & 98 \\
\textit{Stamp Seal} & 95 & 84 & 89 & 77 & 79 & 55 & 93 & 92 & \textbf{96} & \textbf{97} & 90 & 94 \\
\textit{Turn Switch} & 39 & 32 & 54 & 46 & 62 & 54 & \textbf{84} & \textbf{78} & 44 & 45 & 61 & 59 \\
\midrule
\textbf{Average} & \textbf{92.94} & 90.02 & 90.98 & 89.34 & 82.74 & 76.76 & 88.66 & 87.02 & 92.90 & 91.50 & 91.88 & \textbf{91.78} \\
\bottomrule[0.5mm]
\end{tabular}%
}
\end{table*}

\end{document}